\documentclass[review]{elsarticle}

\usepackage{amssymb, amsmath}
\usepackage{booktabs}
\usepackage{multirow}
\usepackage{makecell}
\usepackage{bm}
\usepackage{url}
\usepackage[htt]{hyphenat}
\usepackage{caption}
\usepackage{hyperref}
\hypersetup{colorlinks = true, allcolors = cyan}
\usepackage[capitalise, nameinlink]{cleveref}
\usepackage{color}
\bibpunct{\color{cyan}[}{\color{cyan}]}{,}{n}{}{;}

\journal{Journal of \LaTeX\ Templates}

\begin{document}
\hypersetup{allcolors = cyan}

\begin{frontmatter}

\title{BERT-JAM: Boosting BERT-Enhanced Neural Machine Translation with Joint Attention}

\author{Zhebin Zhang\corref{mycorrespondingauthor}}
\ead{zhebinzhang@zju.edu.cn}

\author{Sai Wu}
\cortext[mycorrespondingauthor]{Corresponding author}
\ead{wusai@zju.edu.cn}

\author{Dawei Jiang}
\ead{jiangdw@zju.edu.cn}

\author{Gang Chen}
\ead{cg@zju.edu.cn}

\address{College of Computer Science and Technology, Zhejiang University, Hangzhou 310007, China}

\begin{abstract}
BERT-enhanced neural machine translation (NMT) aims at leveraging BERT-encoded representations for translation tasks. A recently proposed approach uses attention mechanisms to fuse Transformer's encoder and decoder layers with BERT's last-layer representation and shows enhanced performance. However, their method doesn't allow for the flexible distribution of attention between the BERT representation and the encoder/decoder representation. In this work, we propose a novel BERT-enhanced NMT model called BERT-JAM which improves upon existing models from two aspects: 1) BERT-JAM uses joint-attention modules to allow the encoder/decoder layers to dynamically allocate attention between different representations, and 2) BERT-JAM allows the encoder/decoder layers to make use of BERT's intermediate representations by composing them using a gated linear unit (GLU). We train BERT-JAM with a novel three-phase optimization strategy that progressively unfreezes different components of BERT-JAM. Our experiments show that BERT-JAM achieves SOTA BLEU scores on multiple translation tasks.
\end{abstract}


\end{frontmatter}

% \linenumbers

\section{Introduction}

Pre-training has been demonstrated as a highly effective method for boosting the performance of many natural language processing (NLP) tasks such as question answering, sentimental analysis, and so on. By training on massive unlabeled text data, pre-trained models are able to learn the contextual representations of input words, which are extremely helpful for accomplishing downstream tasks. 
BERT \cite{Delvin19}, as one of the most widely used pre-trained models, is trained using two unsupervised tasks, namely, mask language modeling and next sentence prediction.
By adding a few layers on top, BERT can be easily adapted into a task-specific model, which is then fine-tuned on the labeled data to achieve optimal performance. Such a practice has been exercised in various NLP scenarios and has achieved many state-of-the-art (SOTA) results.

The study of integrating BERT into neural machine translation models, which is referred to as BERT-enhanced NMT, has received much research interest. However, exploiting BERT for NMT is not as straightforward as in other NLP tasks.
The architecture of a typical NMT model consists of an encoder that transforms the source language words into a hidden representation, and a decoder that predicts the target language words based on the hidden representation. The challenge of exploiting BERT for NMT is twofold. Firstly, NMT models are mostly deep neural networks with a parameter size comparable to or even larger than that of BERT, which makes the combined model hard to optimize. Secondly, since existing NMT models are mostly trained with massive samples, the usual practice of fine-tuning BERT on the labeled corpus can lead to the problem of catastrophic forgetting \cite{Goodfellow13, Yang20}.

\begin{figure*}[!t]
    \centering
    \includegraphics[width=\textwidth]{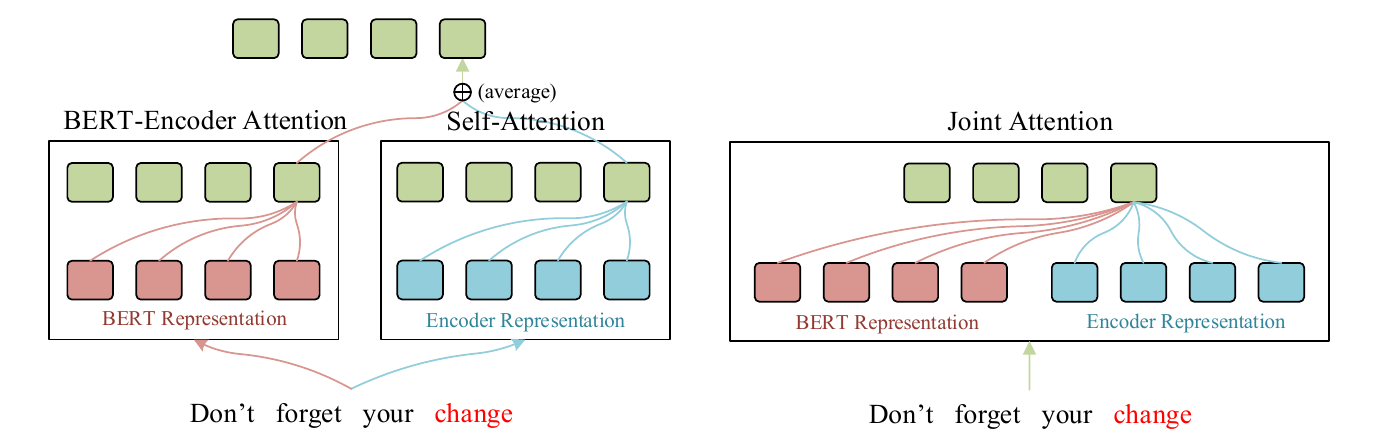}
    \caption{Comparison between using separate attention modules (left) and using the joint module (right).}
    \label{fig:compare}
\end{figure*}

The recently proposed BERT-fused model \cite{Zhu20} uses attention mechanisms to bridge between the NMT model and BERT. For example, they introduce an extra BERT-encoder attention module to fuse the encoder layer with the BERT representation. The outputs of the BERT-encoder attention module and the self-attention module are averaged. 
Consider the case exemplified in \cref{fig:compare} (left), it's more likely that the word \emph{change} should be interpreted as \emph{money} rather than other meanings in this context. However, if the training corpus doesn't contain similar expressions, the model can fail in this translation due to the ambiguity. When BERT representations are introduced, the contextual information learned by BERT can be helpful for the translation. Concretely, a BERT-encoder attention module can be used to capture the pre-trained knowledge embedded in the BERT representation that is absent in the self-attention module. 

However, we find that averaging their outputs means regarding them as equally important, which can hurt performance under some circumstances. In the above example, only the BERT-encoder attention module provides useful information for interpreting the word \emph{change}, while the self-attention module offers faulty or noisy information. Combining their outputs directly can result in confusion during translation. 
Hence we assert that it's essential to allow the model to decide which information to concentrate on.
To this end, we propose to use a joint-attention module to integrate multiple representations that contain different contextual information. As shown in \cref{fig:compare} (right), the learnable weights of the joint-attention module allow it to assign more attention to the BERT representation in this case. Compared with the BERT-fused model, our method is better at augmenting desired information and hence boosts performance.

Although existing BERT-enhanced NMT models mostly focus on leveraging BERT's last-layer representation, we find that the intermediate layers can contain semantic and contextual information that is absent in the last layer and might help improve translation performance.
The dynamic fusion mechanism proposed by \citet{Weng20} allows the Transformer encoder to leverage BERT's intermediate representations. However, their method doesn't work for the decoder at the inference stage because it requires the ground truth as input.
This motivates us to explore feasible techniques for generating composite BERT representations that can be used in both the encoder and the decoder.

In this paper, we introduce a BERT-enhanced NMT model called BERT-JAM, which stands for \textbf{BERT}-fused \textbf{J}oint-\textbf{A}ttention \textbf{M}odel. BERT-JAM is equipped with joint-attention modules that allow the encoder/decoder to selectively concentrate on the BERT representation or the encoder/decoder representation by attending to them simultaneously. Besides, we seek to improve upon the existing BERT-enhanced models by making better use of BERT's intermediate layers. Specifically, we allow each encoder/decoder layer to use a GLU module to transform BERT's intermediate representations into a composite representation used by the joint-attention module.

In order to achieve optimal performance, we train BERT-JAM following a three-phase optimization strategy which progressively unfreezes different components of the model during training. 
We show that fine-tuning BERT is a crucial step to unearth the full potential of BERT-JAM, in contrast to the previous claim that fine-tuning BERT offers few gains \cite{Yang20} for NMT models. Moreover, we study how the BERT-enhanced NMT performance varies with the size of BERT by feeding different BERT models into BERT-JAM, ranging from the most compact BERT with 2 layers and embedding dimension 128 to the standard BERT-base model. This study can be beneficial because it can provide us with a guide on how to adjust the model with minimal performance loss when we have to resort to a smaller model size due to limited computation resources.

We summarize the contributions of this paper as follows: 
\begin{itemize}
   \item We propose a novel BERT-enhance NMT model named BERT-JAM which leverages joint-attention modules for dynamically allocating attention between different representations. Besides, compared with existing approaches, BERT-JAM makes better use of BERT by combining all of its intermediate representations into a composite form through a GLU module.
   \item This is the first work that studies how the size of BERT affects the performance of BERT-enhanced NMT models, and we find out that increasing BERT's embedding dimension rather than its layer number is more crucial to the improvement of translation qualities.
   \item We train BERT-JAM with a novel three-phase optimization strategy that allows us to overcome the catastrophic forgetting problem observed in previous studies.
   \item We evaluate the proposed BERT-JAM model on several widely used translation tasks. Experimental results show that BERT-JAM achieves new SOTA scores on multiple benchmarks, demonstrating the effectiveness of our method.
\end{itemize}

The rest of this paper is organized as follows. 
In \cref{sec:approach}, we introduce our approach to BERT-enhanced NMT where a detailed description of our model will be presented. 
The experimental setups are described in \cref{sec:setup}. 
In \cref{sec:results}, several experiments are conducted and the results are discussed. 
We give a review of related works in \cref{sec:related} and the conclusions are drawn in \cref{sec:conclusion}.

\section{Approach}
\label{sec:approach}
This section presents our proposed approach to boosting machine translation performance with BERT. We begin by introducing some backgrounds of BERT-enhanced NMT. Then, we introduce the construction of the joint-attention module used in our model. Next, we detail the architecture of BERT-JAM. Finally, we describe the three-phase optimization strategy used to train the model.

\subsection{Backgrounds}
\label{ssec:background}

\subsubsection{Neural machine translation}
NMT is modeled as a sequence-to-sequence task that learns the mapping from the source language sentence $X=(x_1, x_2, ..., x_n)$ to the target language sentence $Y=(y_1, y_2, ..., y_m)$.
Existing NMT models are mostly based on the widely adopted encoder-decoder architecture. On the source side, the encoder transforms the source words into a hidden representation
\begin{equation}
    H^E=(h_1^E, h_2^E, ..., h_n^E)=\text{Enc}(X).    
\end{equation}
To calculate the probability of each target word $y_i$, the decoder takes the hidden representation $H^E$ and the previous words $y_{<i}$ as input to obtain the representation for the $i$-th target word as
\begin{equation}
    \label{eq:back_dec}
    h_i^D=\text{Dec}(H^E, y_{<i}).    
\end{equation}
Then a linear projection is applied on $h_i^D$ to map it into the vocabulary size $|V|$, followed by a softmax function. Hence, the probability is given by
\begin{equation}
    \label{eq:back_prob}
    p_{\theta}\left(y_i|H^E, y_{<i}\right)=\text{softmax}(\text{linear}(h_i^D)),
\end{equation}
where $\theta$ represents the parameters of the model. And the probability of the whole sentence is represented as the joint probability of individual words, given by
\begin{equation}
    \label{eq:back_joint_prob}
    p_{\theta}(Y|X)=\prod_{i=1}^m p_{\theta}\left(y_i|H^E, y_{<i}\right).
\end{equation}
The model is optimized by adjusting the weights $\theta$ so that the above probability is maximized.

\subsubsection{BERT-Enhanced NMT} 
To further boost the performance of machine translation, researchers seek to feed NMT models with extra pre-trained knowledge obtained from massive monolingual corpora beyond the limited bilingual data. To this end, BERT-enhanced NMT models have been proposed to make use of the rich contextual information provided by BERT. A common practice to achieve this goal is to use BERT to encode the source language sentence into the representation $H^B=\text{BERT}(X)$, which can be fed into the NMT model in various ways. The straightforward method is to initialize the encoder with BERT, so that the encoder representation $H^E$ in \cref{eq:back_dec,eq:back_prob,eq:back_joint_prob} is replaced by $H^B$.
Another method uses BERT as the embedding layer of the encoder so that the encoder representation becomes $H^E=\text{Enc}(H^B)$. These methods suffer from the problem that the model gradually forgets the pre-learned monolingual information when trained on the bilingual data. Recent studies approach this problem differently. \citet{Yang20} used a distillation method to allow the NMT model to learn from pre-trained representations without forgetting. And \citet{Zhu20} proposed a BERT-fused model that uses attention mechanisms to fuse Transformer's encoder and decoder layers with BERT representations.

\subsection{Joint Attention}
We propose a joint-attention module that transforms a primary vector sequence $R=(r_1, r_2, ..., r_n)$ based on a secondary vector sequence $S=(s_1, s_2, ...,\allowbreak s_m)$. The first step is to project the two sequences of vectors from their original dimensions to a common embedding dimension $d_{model}$. Specifically, we perform three different projections on the primary sequence to respectively derive the query vectors $Q=(q_1, q_2, ..., q_n)$, the primary key vectors $K^R=(k_1^R, k_2^R, ..., k_n^R)$, and the primary value vectors $V^R=(v_1^R, v_2^R, ..., v_n^R)$. And the secondary sequence is projected to obtain the secondary key vectors $K^S=(k_1^S, k_2^S, ..., k_m^S)$ and the secondary value vectors $V^S=(v_1^S, v_2^S, ..., v_m^S)$.

To allow the query vectors to attend to the primary and secondary sequences simultaneously, we concatenate the primary and secondary key/value vectors to obtain the joint key/value vectors, given by
\begin{equation}
    \label{eq:concat}
    \begin{aligned}
        K & = (k_1, k_2, ..., k_{m+n})&= (k_1^R, ..., k_n^R, k_1^S, ..., k_m^S), \\
        V & = (v_1, v_2, ..., v_{m+n})&= (v_1^R, ..., v_n^R, v_1^S, ..., v_m^S).
    \end{aligned}
\end{equation}
Next, we calculate the attention weights between the $i$-th query vector $q_i \in Q$ and the $j$-th key vector $k_j \in K$ as
\begin{equation}
    \label{eq:weights}
    \begin{aligned}
    w_{ij} & = \frac{\exp(e_{ij})}{\sum_{j=1}^{m+n} \exp(e_{ij})}, \\
    e_{ij} & = \frac{q_i \cdot k_j}{\sqrt{d_{model}}}.
    \end{aligned}
\end{equation}
Finally, the output of the joint module is given by
\begin{equation}
    \centering
    \label{eq:joint}
    \begin{aligned}
        R' &  =\text{Attn}_{\text{joint}}(R, S) = (r'_1, r'_2, ..., r'_n), \\
        r'_i & = \sum_{j=1}^{m+n} w_{ij} v_j,
    \end{aligned}
\end{equation}
where $v_j \in V$ represents the $j$-th value vector.

\subsection{BERT-JAM}
\label{ssec:bert-jam}

\begin{figure*}[t]
    \centering
    \includegraphics[width=\textwidth]{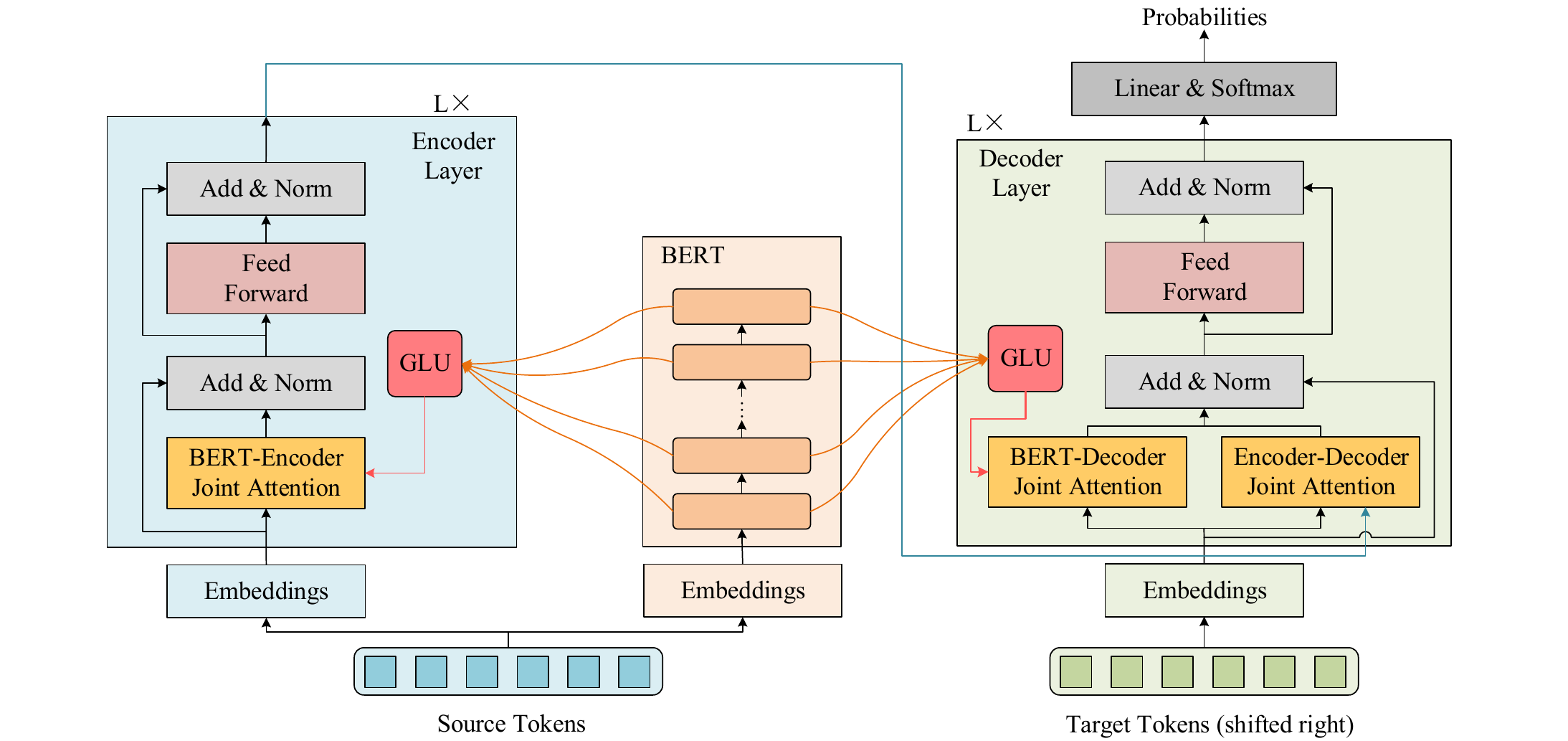}
    \caption{The architecture of BERT-JAM.}
    \label{fig:arch}
\end{figure*}

The architecture of our proposed BERT-JAM is shown in \cref{fig:arch}. BERT-JAM consists of three main parts: a pre-trained BERT model, a multi-layer encoder, and a multi-layer decoder. For any source sequence $X=(x_1, x_2, ..., x_n)$ and the paired target sequence $Y=(y_1, y_2, ..., y_m)$, BERT-JAM processes them following the steps below:

\begin{itemize}

\item \underline{\textit{Step 1:}} 
The source sequence $X$ is first encoded by BERT which consists of $L_B$ layers. Denoting the hidden representation produced by the $i$-th BERT layer as $\bm{B}_i$, we represent the outputs of all BERT layers as $\text{BERT}(X)=(\bm{B}_1, \bm{B}_2, ...., \bm{B}_{L_B})$.

\item \underline{\textit{Step 2:}} 
Each encoder/decoder layer is equipped with an independent GLU module to transform the BERT outputs into a composite representation before fusing it with the encoder/decoder representation. 
The GLU module computes a gate $g$ based on the linear combination of BERT's multi-layer representations, which is used to control the information flowing from the BERT layers into the encoder/decoder layer. Formally, the composite representation $\bm{B}$ is give by
\begin{equation}
    \label{eq:glu}
    \begin{aligned}
    \bm{B} & = g \otimes \left(\sum_{i=1}^{L_B}\alpha_i \bm{B}_i \right), \\
    g & =\sigma\left(\sum_{i=1}^{L_B}\beta_i \bm{B}_i \right),
    \end{aligned}
\end{equation}
where $\{\alpha_i\}$ and $\{\beta_i\}$ are learnable weights, and $\otimes$ is the element-wise product. Note that the weights of the GLU module are not shared across different layers so that each layer can independently compose desired BERT representations.

\item \underline{\textit{Step 3:}}
The encoder contains $L$ identical layers that progressively transform the embedded source sequence of $n$ tokens from lower layers to upper ones. Specifically, the output of the the $i$-th encoder layer, denoted by $\bm{E}^i=(e^i_1, e^i_2, ..., e^i_n)$, serves as the input of the ($i$$+$$1$)-th layer. Note that $\bm{E}^0$ is the output of the embedding layer.
The $i$-th encoder layer obtain the composite BERT representation for the current layer, denoted by $\bm{B}^{E, i}$, by following the second step. Then a BERT-encoder joint-attention module is introduced to transform $\bm{E}^i$ based on $\bm{B}^{E, i}$ as
\begin{equation}
    \label{eq:enc}
    \bm{E}'^i= \text{Attn}_{\text{joint}}(\bm{E}^i, \bm{B}^{E, i}).
\end{equation}
The joint-attention module is followed by a feed-forward network module. Both modules are surrounded by the residual connection \cite{He16} and layer normalization \cite{Ba16}. The output of the top encoder layer is denoted by $\bm{E}^L$.

\item \underline{\textit{Step 4:}}
The decoder also has $L$ layers used to transform the embedded target sequence of $m$ tokens, on the top of which is a linear projection and softmax layer that outputs the probability distribution over the target language vocabulary. Similar to the encoder layer, each decoder layer first obtains the BERT representation $\bm{B}^{D, i}$ for the current layer. The input of this decoder layer, denoted by $\bm{D}^i=(d^i_1, d^i_2, ..., d^i_m)$, is transformed using two different joint-attention modules, including a BERT-decoder joint-attention module and an encoder-decoder joint-attention module. The outputs of the two modules are combined as

\begin{equation}
    \label{eq:dec}
    \bm{D}'^i = \frac{1}{2} (\text{Attn}_{\text{joint}}(\bm{D}^i, \bm{B}^{D, i}) 
    +  \text{Attn}_{\text{joint}}(\bm{D}^i, \bm{E}^L)).
\end{equation}
As in the encoder layer, the decoder layer incorporates a feed-forward network module. The residual connection and layer normalization are applied in a similar way.

\end{itemize}

\subsection{Optimization Strategy}
\label{ssec:training}
Training BERT-enhance NMT models can be trickier than training other BERT-aided models. The main challenge stems from the fact that BERT has a parameter size comparable to that of the encoder and decoder of the NMT model. 
Besides, to obtain a fully trained NMT model, a large amount of bilingual data are generally required. 
Therefore, for a BERT-enhanced NMT model, if we train the whole model all at once, the knowledge pre-learned by BERT can be gradually forgotten after the model has seen enough new training samples. This phenomenon is referred to as catastrophic forgetting \cite{Goodfellow13, Yang20}. 
To cope with this challenge, we train BERT-JAM with a three-phase optimization strategy which gradually unfreezes different parts of the model to achieve its optimal performance. 
Previous studies \cite{Yang20} claim that fine-tuning BERT in NMT models offers no gain. We find this is not the case if we fine-tune BERT at the right time and in a controlled manner, as will be demonstrated in \cref{ssec:ablation}.

Our proposed optimization strategy proceeds in three steps, as described below:
\begin{itemize}
    \item \underline{\textit{Phase 1: Warmup Training.}} 
In this phase, we train BERT-JAM from scratch by keeping the parameters of BERT and the GLU modules frozen. The parameters of the encoder and the decoder are fully trained before we proceed to the next phase.
It is noteworthy that we initialize each GLU module as
\begin{equation}
\begin{aligned}
    \alpha_i  &= \left\{
        \begin{aligned}
            1, \quad & i = L_B \\
            0, \quad & i \in [1, L_B-1]
        \end{aligned}
    \right. \\
    \beta_i  &=0, \quad i \in [1, L_B]. 
\end{aligned}
\end{equation}
Replacing $a_i$ and $b_i$ in \cref{eq:glu} with the above initialization yields $g=\frac{1}{2}$ and $\bm{B}=\frac{\bm{B}_{L_B}}{2}$, which means that we initialize BERT-JAM so that only the last layer of BERT is fed into the encoder and the decoder. In order to compensate for the halved BERT representation, we multiply the GLU output by $2$ only in this phase. 
The reason for this initialization is that the way BERT is trained determines that its last-layer representation contains the most useful information for downstream tasks. 
And starting from the most salient representation allows the model to converge faster.

\item \underline{\textit{Phase 2: Adjust GLU weights.}} 
After the parameters in the encoder and the decoder are fully trained, we unfreeze the GLU modules contained in each encoder/decoder layer to allow BERT's intermediate layers to contribute to the model. As the training progresses, each GLU module gradually learns the optimal weights used for combining BERT's intermediate representations into a well-formed composition. 

\item \underline{\textit{Phase 3: Fine-tune BERT.}} 
In the final phase, we unfreeze and fine-tune BERT to further improve the performance. Due to the catastrophic forgetting problem, it's crucial that we don't overfit the model on the training set. With this in mind, we keep a careful eye on the validation loss at the end of each epoch and stop training when we observe degraded performance.

\end{itemize}

We justify our three-phase optimization strategy with the experimental results to be presented in \cref{ssec:ablation} where we show that skipping any of the above steps can result in performance loss.

\section{Experimental Setup}
\label{sec:setup}
This section details the experimental setup in terms of data preparation, model configuration, training settings, and evaluation metrics. We implement our model upon the Fairseq repository \footnote{\url{https://github.com/pytorch/fairseq}}.
We will make the code and other precessing scripts publicly available.

\subsection{Data Preparation}
We conduct extensive experiments on multiple translation tasks in both low-resource and high-resource scenarios. For the low-resource scenario, we work on five IWSLT tasks where each training corpus contains a few hundred thousand sentence pairs. For the high-resource scenario, we evaluate on the WMT'14 En-De dataset which contains millions of sentence pairs. We detail the processing of the datasets as follows.

\subsubsection{IWSLT}
For the low-resource scenario, we evaluate BERT-JAM on five IWSLT translation tasks, namely, IWSLT'14 English-German (En-De), IWSLT'14 German-English (De-En), IWSLT'14 English-Spanish (En-Es), IWSLT'17 English-French (En-Fr) and IWSLT'17 English-Chinese (En-Zh). 
We follow the setup in \cite{Zhu20} to pre-process the datasets.
Specifically, letters are lowercased for the En-De and De-En tasks only. And words in all datasets are tokenized using Moses toolkit \cite{Koehn07} and split into sub-words using byte pair encoding (BPE) \cite{Sennrich16} with $10k$ symbols. A joined vocabulary is built by merging source and target sentences for each language pair. For the En-De and De-En tasks, the dataset contains $160k$ training examples with $7k$ of them drew out for validation. And the concatenation of \emph{dev2010, dev2012, tst2010, tst2011, tst2012} is used as the test set. For the En-Es, En-Fr and En-Zh tasks, the training sets contain $183k$, $236k$, and $235k$ sentence pairs respectively and the TED Talk files of the corresponding years are pre-processed as validation/testing sets. 

\subsubsection{WMT}
For the high-resource scenario, we evaluate BERT-JAM on the WMT'14 En-De dataset in both translation directions which contains$4.5M$ training sentence pairs. We tokenize the data as in the IWSLT tasks without lowercasing the letters. Also, words are split into sub-words using BPE with $32k$ symbols and a joined vocabulary is built. We develop on the concatenation of \emph{newstest2012} and \emph{newstest2013} and test on \emph{newstest2014}.

\subsection{Model Configuration}
For all the translation tasks, we equip BERT-JAM with 6 encoder layers and 6 decoder layers. The dropout ratio is set to 0.3.
Denote by $d_{model}$, $d_{ff}$ and $N_{head}$ the embedding dimension, the feed-forward network dimension and the number of attention heads respectively. For the IWSLT tasks, we set $d_{model}=512$, $d_{ff}=1024$ and $N_{head}=4$. For the WMT tasks, we employ a larger model with $d_{model}=1024$, $d_{ff}=4096$ and $N_{head}=16$.

Depending on the translation tasks, we equip BERT-JAM with BERT models of different sizes pre-trained for different languages which are publicly available, as listed in \cref{tab:bert_model}. For IWSLT tasks we choose $\text{BERT}_{\text{BASE}}$ with 12 encoder layers and embedding dimension 768. Note that the \texttt{bert-base-german-uncased}\footnote{\url{https://s3.amazonaws.com/models.huggingface.co/bert/bert-base-german-dbmdz-uncased-pytorch_model.bin}} and \texttt{bert-base-uncased}\footnote{\url{https://s3.amazonaws.com/models.huggingface.co/bert/bert-base-uncased.tar.gz}} models are pre-trained on German and English corpora respectively.
We have intended to use the $\text{BERT}_{\text{LARGE}}$ model with 24 encoder layers and embedding dimension 1024 for the WMT tasks. But since there's no available $\text{BERT}_{\text{LARGE}}$ pre-trained for German, we resort to the smaller \texttt{bert-base-german-uncased} for the WMT De-En task. As for the WMT En-De task, we use \texttt{bert-large-uncased}\footnote{\url{https://s3.amazonaws.com/models.huggingface.co/bert/bert-large-uncased.tar.gz}} which is the enlarged version of \texttt{bert-base-uncased}.

As part of our experiments to be presented in \cref{ssec:vary_size}, we will evaluate BERT-JAM on IWSLT'14 En-De with pre-trained BERT models of varying sizes ranging from the most compact BERT model with 2 encoder layers and embedding dimension 128 to the largest one which is the same as $\text{BERT}_{\text{BASE}}$. These pre-trained models are available at this repository\footnote{\url{https://github.com/google-research/bert}}.

\begin{table}[ht]
    \caption{BERT models used for different tasks.}
    \label{tab:bert_model}
    \centering
    \begin{tabular}{c|c|c}
        \toprule
        \multicolumn{2}{c|}{\textbf{Task}} & \textbf{BERT model} \\
        \midrule
        \multirow{5}{*}{IWSLT} & De-En & \texttt{bert-base-german-uncased} \\
        \cline{2-3}
        & En-De & \multirow{4}{*}{\texttt{bert-base-uncased}} \\
        & En-Es \\
        & En-Fr \\
        & En-Zh \\
        \hline
        \multirow{2}{*}{WMT} & De-En & \texttt{bert-base-german-uncased} \\
        \cline{2-3}
        &  En-De & \texttt{bert-large-uncased}\\
        \bottomrule
    \end{tabular}
\end{table}

% https://s3.amazonaws.com/models.huggingface.co/bert/bert-base-german-dbmdz-uncased-vocab.txt
% https://s3.amazonaws.com/models.huggingface.co/bert/bert-base-german-dbmdz-uncased-config.json

\subsection{Training}
All the experiments are conducted on a single machine equipped with 4 NVIDIA V100 GPUs. We train the model using the Adam optimizer \cite{Kingma15} with $\beta_1=0.9$ and $\beta_2=0.98$. The learning rate follows the inverse square root schedule which is firstly linearly warmed up from $10^{-7}$ to $0.0005$ for $4k$ steps and then given by
\begin{equation}
    lr=0.0005\frac{\sqrt{step_{warm\_up}}}{\sqrt{step_{current}}}.
\end{equation}
The models are trained in batches containing up to $4k$ tokens. For the WMT tasks, we accumulate the gradients for 32 iterations before updating to simulate training on 128 GPUs.
For all tasks, we follow the three-phase optimization strategy described in \cref{ssec:training}. And we obtain the model used for testing by averaging the checkpoint weights of the last ten epochs. We train the model until convergence in the first two phases. In the third phase, since too much training makes the pre-trained knowledge become forgotten, we stop when the averaged checkpoint achieves minimal loss on the validation set, as will be further explained in \cref{ssec:ablation}.

\subsection{Evaluation}

We use beam search to generate target sentences on the test sets in inference mode.
Beam search is a heuristic search algorithm widely used for sequence-to-sequence generation. Instead of selecting the word with the maximum probability at each decoding step as in the greedy search algorithm, beam search tries to find the most likely sequence of words based on the joint probability. It is parameterized by a beam width which is the size of the candidate set. Besides, length penalty is applied to handle sentences of different lengths.
In our experiments, for the IWSLT tasks, we set beam width to 5 and length penalty to 1. As for the WMT tasks, the beam width is 4 and the length penalty is 0.6.

BLEU score \cite{Papineni02} is a commonly used metric for the measurement of translation qualities. 
It works by counting matching n-grams between the candidate translation and the reference text. 
However, since BLEU is a parameterized metric, different choices of the parameters can lead to the variation of the scores. For a fair comparison with existing studies, we follow the same implementations of BLEU described in their papers. Specifically, for translation tasks in both directions on the IWSLT'14 En-De and the WMT'14 En-De datasets, we use \texttt{multi-bleu.perl}\footnote{https://github.com/moses-smt/mosesdecoder/blob/master/scripts/generic/multi-bleu.perl}. For other tasks, we use detokenized \textsc{SacreBLEU} \cite{Post18} instead.
Note that for the WMT'14 En-De task only, we additionally perform compound splitting following \citet{Vaswani17} before calculating BLEU scores to produce comparable results.

\section{Experiments and Results}
\label{sec:results}
This section introduces the experiments we conduct to evaluate our model. We first report the results on the benchmark translation tasks in both low-resource and high-resource scenarios. Then we work on the IWSLT'14 En-De task where we vary the size of BERT to study how it affects the translation performance. Finally, we conduct ablation studies to justify the design choices in our proposed model.

\subsection{Main Results}

\subsubsection{IWSLT}
For the low-resource scenario, we experiment on multiple IWSLT datasets. 
We present in \cref{tab:iwslt_de2en} the results on the widely used IWSLT'14 De-En dataset. As shown, BERT-JAM outperforms the previous SOTA model by a large margin, setting a new SOTA score of 38.66 on this task.

\begin{table}[ht]
    \caption{Translation qualities on the IWSLT'14 De-En task.}
    \label{tab:iwslt_de2en}
    \centering
    \begin{tabular}{lcc}
        \toprule
        \textbf{Model}            & \textbf{BLEU}    \\
        \midrule
        Transformer \cite{Vaswani17}  &   34.64  \\
        DynamicConv \cite{Wu19}         &     35.2  \\
        BERT-fused \cite{Zhu20} & 36.11 \\
        MAT \cite{Fan20} & 36.22 \\
        MUSE \cite{Zhao19} & 36.3 \\
        MAT+Knee \cite{Iyer20} & 36.6   \\
        BERT-JAM   &   \textbf{38.66}  \\
        \bottomrule
    \end{tabular}
\end{table}

The results of the IWSLT En-X (X$\in$\{De, Es, Fr, Zh\}) tasks are presented in \cref{tab:iwslt_en2x}. We compare BERT-JAM with two baseline models, the Transformer and the BERT-fused model.
As shown, BERT-JAM outperforms the BERT-fused model on all translation tasks except En-Zh. The findings could be explained by the hypothesis that our proposed joint-attention module is better at translating between languages that share cognate words. As for distant language pairs such as English and Chinese, attending to their concatenated token representations can hurt the performance.

\begin{table}[ht]
    \caption{Translation qualities on the IWSLT En-X tasks.}
    \label{tab:iwslt_en2x}
    \centering
    \begin{tabular}{cccccc}
        \toprule
        & Transformer-Base & BERT-fused &  BERT-JAM    \\         
        \midrule
        En-De  &   28.57  &   30.34  &   \textbf{31.20} \\
        En-Es  &   39.0   &   41.4   &   \textbf{42.3}  \\
        En-Fr  &   35.9   &   38.7   &   \textbf{39.8}  \\
        En-Zh  &   26.3   &   \textbf{28.2}  &   27.9   \\
        \bottomrule
    \end{tabular}
\end{table}

\subsubsection{WMT}
For the high-resource scenario, we experiment on the WMT'14 En-De and De-En tasks. 
The WMT'14 En-De dataset is one of the most used benchmark tasks for evaluating NMT models. As reported in \cref{tab:wmt_en2de}, we compare BERT-JAM with strong baseline models including those which make use of pre-trained models (the lower half) and those which don't (the upper half). BERT-JAM outperforms all of the previous models and achieves a BLEU score of 31.59. Our result is only inferior to those models that utilize extra data for training \cite{Edunov18, Raffel19}.

\begin{table}[ht]
    \caption{Translation qualities on the WMT'14 En-De task.}
    \label{tab:wmt_en2de}
    \centering
    \begin{tabular}{lcc}
        \toprule
        \textbf{Model}            & \textbf{BLEU}    \\
        \midrule
        Transformer \cite{Vaswani17}  &   29.3  \\
        DynamicConv \cite{Wu19}         &     29.7  \\
        Evolved Transformer \cite{So19}          &   29.8    \\
        MUSE \cite{Zhao19} & 29.9 \\
        \midrule
        \citet{Imamura19} & 29.04 \\
        \textsc{Apt} framework \cite{Weng20} & 29.23 \\
        \textsc{CTnmt} \cite{Yang20} &   30.1   \\
        BERT-fused \cite{Zhu20}      &   30.75  \\
        BERT-JAM                     &   \textbf{31.59}  \\
        \bottomrule
    \end{tabular}
\end{table}

For the WMT'14 De-En task, as shown in \cref{tab:wmt_de2en}, by comparing to the latest results reported in the literature so far, our BERT-JAM achieves a BLEU score of 33.85, advancing the SOTA score by 2 points.

\begin{table}[ht]
    \caption{Translation qualities on the WMT'14 De-En task.}
    \label{tab:wmt_de2en}
    \centering
    \begin{tabular}{lcc}
        \toprule
        \textbf{Model}            & \textbf{BLEU}    \\
        \midrule
        FlowSeq-large \cite{Ma19} & 28.29  \\
        Transformer \cite{Vaswani17} & 31.44 \\
        MAT+Knee \cite{Iyer20} & 31.9   \\
        BERT-JAM   &   \textbf{33.85}  \\
        \bottomrule
    \end{tabular}
\end{table}

\subsection{Varying BERT Size}
\label{ssec:vary_size}
We work on IWSLT'14 En-De to explore the effect of the size of the BERT model on the performance of BERT-enhanced translation. We download pre-trained BERT models of different sizes and incorporate them into BERT-JAM. All models are fully trained following the three-phase strategy. The results are presented in \cref{tab:vary_size} where $L$ and $H$ stand for the number of layers and the embedding dimension of BERT respectively. An intuitive conclusion comes naturally that larger BERT models are better at assisting translation tasks than the more compact ones. Taking a closer at the figures provides us with more insights. 

By comparing the horizontal changes in the scores with the vertical ones, we find out that more performance gains can be achieved by adopting a higher embedding dimension than by deepening the model with more BERT layers. For example, for the BERT model with $L2/H128$, doubling the layer number ($L4/H128$) yields a score of 28.77, while doubling the embedding dimension ($L2/H256$) yields a higher score of 29.18. This is a general pattern that can be verified throughout the table. And it enlightens us that, when we have to resort to a more compact model due to limited computation resources, we should prioritize the embedding dimension of BERT over the layer number.

\begin{figure}[ht]
    \centering
    \includegraphics[width=\columnwidth]{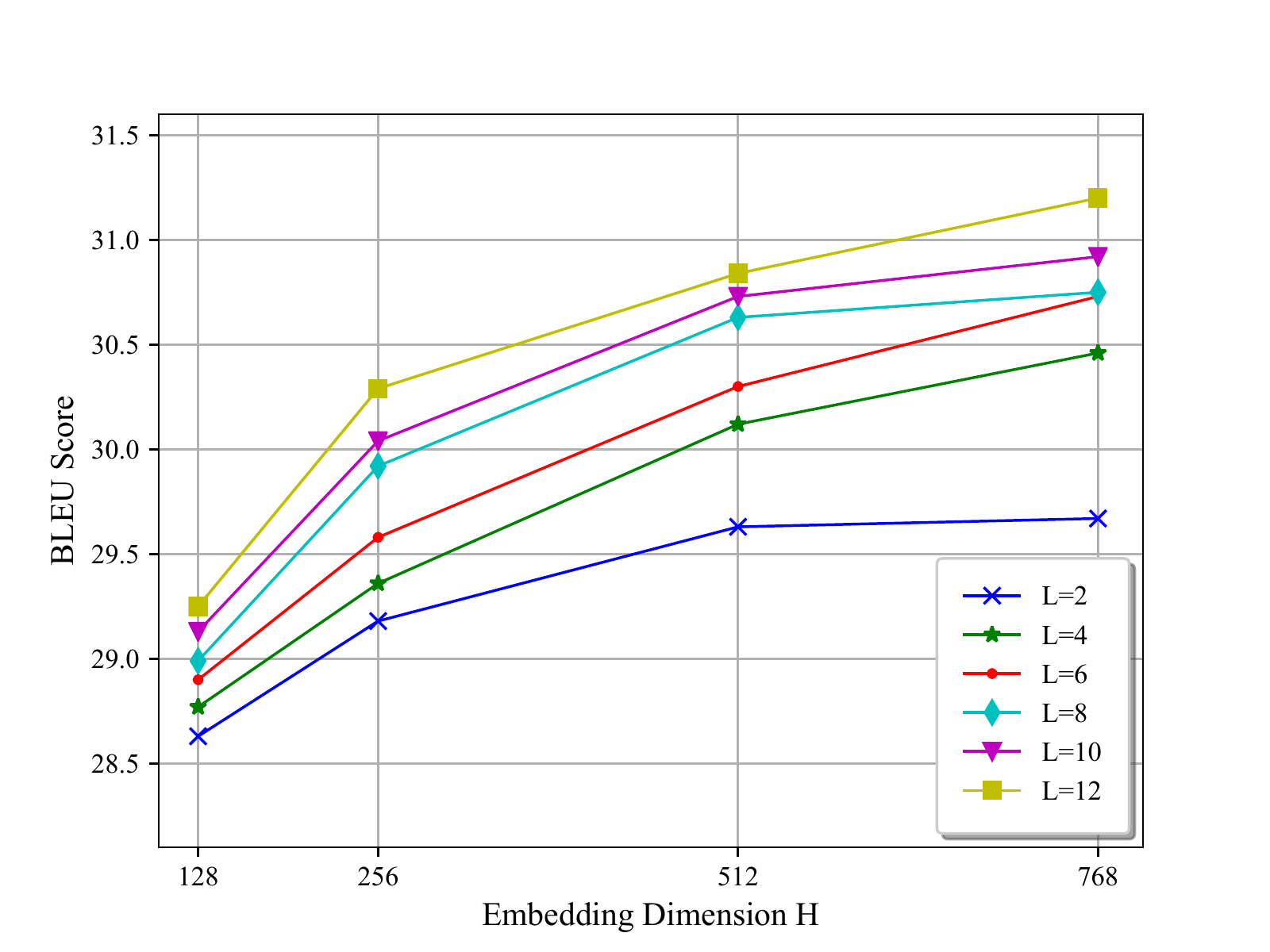}
    \caption{Plot of BLEU scores with varying embedding dimensions of BERT.}
    \label{fig:vary_size}
\end{figure}

However, we cannot expect a steady improvement in translation performance by indefinitely increasing the embedding dimension.
\cref{fig:vary_size} plots the curves of BLEU scores when we vary the embedding dimension of BERT with different layer numbers.
All plotted curves share the same characteristic that the slopes decrease as the dimension increases, indicating that the performance gains can be marginalized at a certain stage where the dimension becomes high enough.

\begin{table}[ht]
    \caption{Translation qualities on the IWSLT'14 En-De task with varying BERT sizes.}
    \label{tab:vary_size}
    \centering
    \begin{tabular}{lcccc}
        \toprule
                    & $H=128$ & $H=256$ & $H=512$ & $H=768$ \\
        \midrule
        $L=2$   &   28.63   &   29.18   &   29.63   &   29.67   \\
        $L=4$   &   28.77   &   29.36   &   30.12   &   30.46   \\
        $L=6$   &   28.90   &   29.58   &   30.30   &   30.73   \\
        $L=8$   &   28.99   &   29.92   &   30.63   &   30.75   \\
        $L=10$  &   29.13   &   30.04   &   30.73   &   30.92   \\
        $L=12$  &   29.25   &   30.29   &   30.84   &   31.20   \\
        \bottomrule
    \end{tabular}
\end{table}

\subsection{Ablation Study}
\label{ssec:ablation}

In order to justify the design choices we make for the model architecture and the training strategy, we conduct a group of ablation studies on the IWSLT'14 De-En task. To this end, we compare the full-featured BERT-JAM with different variations. We report the BLEU scores of these models after finishing each optimization phase, as shown in \cref{tab:ablation}.

\begin{table}[ht]
    \caption{Translation qualities on the IWSLT'14 De-En task in terms of BLEU scores for each variation at the end of each phase.}
    \label{tab:ablation}
    \centering
    \begin{tabular}{l|c|c|c}
        \toprule
        \multirow{2}[5]{*}{\textbf{Model}}  &   \multicolumn{3}{c}{\textbf{BLEU score}} \\
        \cline{2-4} & \makecell{phase 1 \\(40 epochs)} & \makecell{phase 2 \\ (10 epochs)} & \makecell{phase 3 \\ (10 epochs)} \\
        \midrule
        $\mathcal{M}_0$ & 36.85 & 37.25 & 38.66 \\
        % \hline
        $\mathcal{M}_1$ & 36.71 & 37.07 & 38.39 \\
        % \hline
        $\mathcal{M}_2$ & \multicolumn{2}{c|}{36.97} &  38.35 \\
        % \hline
        $\mathcal{M}_3$ & \multicolumn{2}{c|}{37.12} & 38.17 \\
        \bottomrule
    \end{tabular}
\end{table}

\begin{figure}[ht]
    \centering
    \includegraphics[width=\columnwidth]{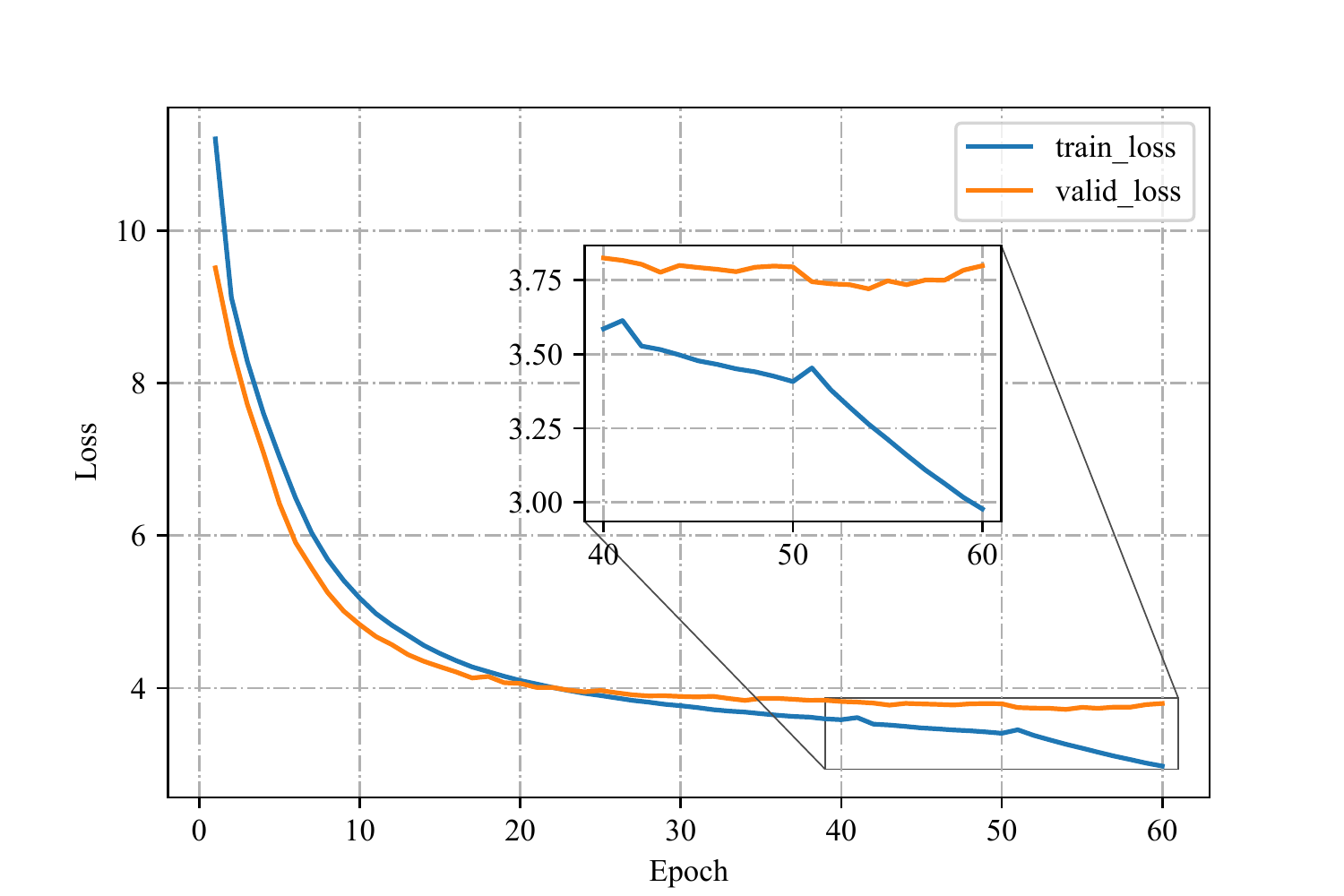}
    \caption{Training and validation loss on the IWSLT'14 De-En task as a function of training epochs.}
    \label{fig:loss}
\end{figure}

\begin{itemize}
    \item $\mathcal{M}_0$: We first report the scores of the full-featured BERT-JAM after each optimization phase. Note that at the end of the first phase where only the last-layer representation of BERT is used, BERT-JAM has already achieved a BLEU score of 36.85, surpassing the previous score of 36.11 reported for the BERT-fused model \cite{Zhu20}. This demonstrates that our proposed joint-attention module makes better use of BERT for the translation than their method. 
    We plot the training and validation loss in \cref{fig:loss} to give you an intuitive understanding of the training process. Notably, the fine-tuning step (epoch 41$\sim$50) makes the validation loss to show a clear decline at first and then rebound. The decline can be explained by the further optimization brought by the adjustment of BERT's parameter, while the rebound is due to the fact that too much training makes BERT forget the pre-learned knowledge. The precipitous drop of the training loss confirms the overfitting of the model on the training set. To take advantage of the benefit of fine-tuning while avoiding the downside, we stop training when the lowest point of the validation loss falls in the middle of the averaging interval (recall that we average the weights of the last ten epochs to obtain the model used for testing). 
    As shown in \cref{tab:ablation}, this fine-tuning strategy allows us the bumps up the BLEU score from 37.25 to 38.66. 

    \item $\mathcal{M}_1$: We substitute the GLU module in BERT-JAM with a linear projection which simply takes a weighted sum of BERT's intermediate representations. The three-phase training strategy is used where we train the projection weights instead in the second phase. By comparing this variation with the full-featured BERT-JAM, we observe that the GLU module boosts the performance from 38.39 to 38.66.
    
    \item $\mathcal{M}_2$: We cancel the GLU module in BERT-JAM and only use the last-layer representation of BERT. We skip the second phase during training and prolong the first phase for ten epochs. As a result, the BLEU score drops from 38.66 to 38.35, showing that using BERT's intermediate representations as supplements does boost the translation performance.
    
    \item $\mathcal{M}_3$: We skip the warmup training and start directly from the second phase where we train all module weights altogether except BERT for 50 epochs, followed by fine-tuning BERT for ten epochs. The resultant score drops significantly from 38.66 to 38.17, showing that the warmup training is crucial for preparing the encoder and the decoder before introducing BERT's intermediate representations.
\end{itemize}

\section{Related Work}
\label{sec:related}
\subsection{Neural Machine Translation} 
Neural machine translation (NMT), which aims to use artificial neural networks for the translation between human languages, has drawn continuous research attention over the past decade. 
NMT is proposed in contrast to traditional statistical machine translation (SMT) such as phrase-based SMT. Instead of trying to tune many sub-components which requires heavy engineering, NMT uses a simpler end-to-end model to generate translations word by word. The main difference that distinguishes an NMT model from traditional SMT models is that it uses continuous vector space for the representation of words \cite{Kalchbrenner13}. And the continuous representations allow the NMT model to efficiently learn the mapping between source and target sentences without suffering from the problem of sparsity. It has been demonstrated that even a naive and straightforward NMT model can outperform the widely studied and mature SMT systems \cite{Sutskever14}.

Most of the NMT models used today are based on the encoder-decoder architecture which was first proposed by \citet{Cho14a} formally. 
At that time, most NMT models were based on recurrent neural networks (RNN).
\citet{Cho14a} first used an RNN encoder to transform the source sentence into a fixed-length context vector and then used an RNN decoder to predict the target sentence based on the vector. 
This practice has been followed by later researches \cite{Luong15a, Cho14b}.

After realizing that the fixed-length vector is the bottleneck of the NMT model, \citet{Bahdanau14} proposed an attention mechanism that allows the decoder to concentrate on certain parts of the source sentence when translating. 
Many of the subsequent studies on NMT are under the same framework of an encoder-decoder architecture and an attention-based approach. 
The key feature that distinguishes these studies from each other is the block-building structures used in their models. Some of them focused on the recurrent structures including the standard RNN \cite{Bahdanau14}, LSTM \cite{Wu16, Luong15b}, and GRU \cite{Jean15}. Others explored replacing the recurrent units with convolutional structures \cite{Gehring17, Wu19}. 
\citet{Vaswani17} took the idea of attention one step further and proposed the famous Transformer model. The transformer layers dispense with any recurrent or convolutional units and depend solely on the attention mechanism to perform the transformation of hidden states. Transformer Follow-up researches are devoted to improving Transformer with various techniques \cite{Gu19, Fonollosa19, Indurthi19}.

\subsection{BERT-Enhanced NMT}
BERT \cite{Delvin19} is essentially a Transformer encoder that is pre-trained on two tasks, namely, masked language modeling and next sentence prediction. It takes as input a sequence of words and encodes them into hidden representations.
A large amount of unlabeled text data are used for the pre-training such that BERT can learn contextualized representations of words. 
After encoded by BERT, the input sequences are associated with rich contextual information which greatly assists natural language understanding and natural language generation tasks. It has been shown that building neural models upon BERT is highly effective in achieving significant performance gains for such tasks.

BERT-enhanced NMT aims at improving translation performance by utilizing the BERT representations. Researchers take different approaches to exploiting BERT for NMT. 
\citet{Imamura19} proposed to substitute the Transformer encoder with a pre-trained BERT and optimize on the bilingual corpus. 
In order to cope with the problem of catastrophic forgetting, \citet{Yang20} proposed an asymptotic distillation method to transfer the pre-trained information from BERT to the NMT model.
\citet{Clinchant19} performed a systematic study of different approaches to boosting translation performance using BERT.
After comparing several existing BERT-enhanced NMT approaches, \citet{Zhu20} proposed a method that uses attention mechanisms to fuse the Transformer layers with the BERT representation.

Our work is partly based on the idea of attention proposed by \citet{Zhu20}. What distinguishes our method from theirs is that we propose a joint-attention module used for the fusion of multiple representations and that we combine all of BERT's intermediate representations using a GLU module. 
Our joint attention module is similar to the mixed attention used in a previous work by \citet{He18}.
They employ mixed attention to allow the encoder and the decoder to share layer-wise features while we build joint attention to integrate multiple different representations.
Besides, the idea of making full use of BERT's intermediate representations has been explored in the previous work by \citet{Weng20} who used a dynamic fusion method to combine these representations. However, their method applies only on the encoder side because it would require the ground truth as input on the decoder side, which is infeasible due to the lack of such knowledge during inference. While our method doesn't suffer from such a restriction and allows both the encoder and the decoder to exploit BERT's intermediate representations.

\section{Conclusion}
\label{sec:conclusion}
In this work, we propose a BERT-enhanced NMT model called BERT-JAM which uses joint-attention to incorporate BERT representations into the NMT model. In contrast to existing models that only utilize BERT's last-layer representation, we make full use of BERT's intermediate representations by composing them through a GLU module. Ablation studies demonstrate that feeding BERT's intermediate representations into the NMT model does improve translation qualities. 
Besides, we adopt a novel three-phase optimization strategy for training the model to overcome the catastrophic forgetting problem found by previous studies in the course of fine-tuning BERT for NMT models. We show that fine-tuning BERT as the last optimization step is beneficial to further boost the performance, but it's crucial that it is under control so that the model doesn't overfit on the training data.
Additionally, by studying the impact of the size of BERT on the performance of BERT-enhanced NMT models, we find that increasing the embedding dimension of BERT, rather than its layer number, is a more cost-effective way to obtain performance gains.
The comprehensive evaluation shows that BERT-JAM outperforms existing models and achieves new SOTA BLEU scores on multiple translation tasks, demonstrating the effectiveness of our method. 

As in many previous works, our method of incorporating BERT into an NMT model introduces extra parameters and makes the training and inference time longer than a regular NMT model. 
Although previous works have explored substituting the encoder of the NMT models with BERT to keep down the model size \cite{Imamura19}, such an approach has been shown to achieve limited performance gains \cite{Zhu20}.
These trade-offs between performance and speed require further studies.
We leave it as our future work to address this downside and study how to leverage BERT for NMT models while considering both speed and translation qualities.

\section*{Acknowledgements}
This work was supported by the Key Research and Development Program of Zhejiang Province of China (Grant No. 2020C01024); the Natural Science Foundation of Zhejiang Province of China (Grant No. LY18F020005); and the National Natural Science Foundation of China (Grant Nos. 61872315, 61672455).

\bibliography{reference.bib}

\end{document}